\documentclass[11pt]{article}
\usepackage{acl2012}
\usepackage{epsfig}
\usepackage{times}
\usepackage{latexsym}
\usepackage{tikz}
\usepackage{multirow}
\usepackage{amsmath,amssymb}
\newcommand{\remove}[1]{}

\newcommand{\tf}{\text{\em tf}}

\newtheorem{lemma}{Lemma}
\newcommand{\qedsymb}{\hfill{\rule{2mm}{2mm}}}
\newenvironment{proof}{\begin{trivlist}
\item[\hspace{\labelsep}{\bf\noindent Proof: }]
}{\qedsymb\end{trivlist}}

\newcommand{\squishlist}{
 \begin{list}{$\bullet$}
  { \setlength{\itemsep}{0pt}
     \setlength{\parsep}{3pt}
     \setlength{\topsep}{3pt}
     \setlength{\partopsep}{0pt}
     \setlength{\leftmargin}{1.5em}
     \setlength{\labelwidth}{1em}
     \setlength{\labelsep}{0.5em} } }

\newcommand{\squishlisttwo}{
 \begin{list}{$\bullet$}
  { \setlength{\itemsep}{0pt}
     \setlength{\parsep}{0pt}
    \setlength{\topsep}{0pt}
    \setlength{\partopsep}{0pt}
    \setlength{\leftmargin}{1em}
    \setlength{\labelwidth}{1.5em}
    \setlength{\labelsep}{0.5em} } }

\newcommand{\squishend}{
  \end{list}  }

\title{Estimating Confusions in the ASR Channel for\\Improved Topic-based Language Model Adaptation}

\author{Damianos Karakos\\
Raytheon BBN Technologies\\
Cambridge, MA\\
{\tt dkarakos@bbn.com}
	  \And
	Mark Dredze \and Sanjeev Khudanpur\\
Human Language Technology Center of Excellence\\
Center for Language and Speech Processing\\
Johns Hopkins University\\
Baltimore, MD, 21211\\
  {\tt mdredze,khudanpur@jhu.edu}}
\date{}

\begin{document}

\begin{titlepage}

\begin{center}

\textsc{\LARGE The Johns Hopkins University}\\[0.5cm]


\includegraphics[width=0.4\textwidth]{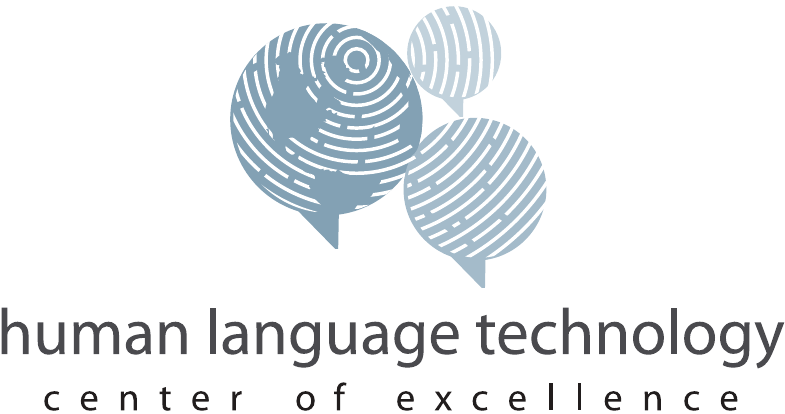}\\[1cm]    

\rule{\linewidth}{0.5mm} \\[0.4cm]
{ \huge \bfseries Estimating Confusions in the ASR Channel for\\Improved Topic-based Language Model Adaptation}\\[0.4cm]

\rule{\linewidth}{0.5mm} \\[0.5cm]

{\Large {\bf Damianos Karakos, Mark Dredze, Sanjeev Khudanpur}}\\[0.9cm]

\textsc{\Large Technical Report 8}\\[0.5cm]
\textsc{\Large \today}

\vfill

\end{center}

\pagebreak

{\Large {\bf \copyright HLTCOE, 2013}}\\[8.0cm]

{\Large \noindent {\bf Acknowledgements} This work is supported, in part, by the Human Language Technology Center of Excellence. Any opinions, findings, and conclusions or recommendations expressed in this material are those of the authors and do not necessarily reflect the views of the sponsor.}\\

\vfill

{\Large \noindent HLTCOE\\ 810 Wyman Park Drive\\ Baltimore, Maryland
  21211\\ http://hltcoe.jhu.edu}

\end{titlepage}

\maketitle
\begin{abstract}
Human language is a combination of elemental languages/domains/styles that change across 
and sometimes within discourses.
Language models, which play a crucial role in speech recognizers and machine translation systems, are particularly
sensitive to such changes, unless some form of adaptation takes place. One approach to speech language model adaptation is self-training,
in which a 
language model's parameters are tuned based on automatically transcribed audio. However, transcription errors can 
misguide self-training, particularly 
in challenging settings such as conversational speech. In this work, we propose a model that 
considers the confusions (errors) of the ASR channel. By modeling the likely confusions in the ASR output instead of using just the 1-best,
we improve self-training efficacy by obtaining a more reliable reference transcription estimate. We demonstrate improved
topic-based language modeling adaptation results over both 1-best and lattice self-training using our ASR channel confusion estimates on telephone 
conversations.
\end{abstract}

\section{Introduction}
Modern statistical automatic speech recognition (ASR) systems rely on language models for ranking hypotheses generated
from acoustic data. Language models are trained on millions of words (or more) taken from text that matches
the spoken language, domain and style of interest. Reliance on (static) training data makes language models brittle \cite{Bellegarda01}
to changes in domain.
However, for many problems of interest, there are numerous hours of spoken audio but little to no written
text for language model training.
In these settings, we must rely on language model adaptation using the spoken audio to improve performance on data of interest.
One common approach to language model adaptation is self-training \cite{nsm09}, in which the language model is retrained on
the output from the ASR system run on the new audio. Unfortunately, self-training learns from both
correct and error ridden transcriptions that mislead the system, a particular problem for high word error rate (WER) domains,
such as conversational speech.
Even efforts to consider the entire ASR lattice in self-training cannot account for the ASR
error bias. Worse still, this is particularly a problem for rare content words as compared to common function words;
the fact that content words are more important for understandability exacerbates the problem.

Confusions in the ASR output pose problems for other applications,
such as speech topic classification
and spoken document retrieval.
In high-WER scenarios, their performance degrades, sometimes considerably
\cite{HR2008}.

In this work, we consider the problem of topic adaptation of a speech recognizer \cite{SR97}, in which we adapt the language model
to the topic of the new speech. Our novelty lies in the fact that we correct for the biases present in the output
by estimating ASR confusions.
Topic proportions are estimated via a probabilistic graphical model which
accounts for confusions in the transcript and provides a more accurate portrayal of the spoken audio. To demonstrate the
utility of our model, we show improved results over traditional self-training as well as lattice based self-training for
the challenging setting of conversational speech transcription. In particular, we show statistically significant
improvements for content words.

Note that \newcite{BRS04} also consider the problem of language model adaptation as an error correction problem, but with
{\em supervised} methods. They train
an error-correcting perceptron model
on reference transcriptions from the new domain.
In contrast, our approach does not assume the
existence of transcribed data for training a confusion model; rather, the model is trained in an unsupervised manner based only on
the ASR output.

The paper proceeds as follows: Section \ref{sec:problem_formulation} describes our setting of language model adaptation and our topic based language model. Section 
\ref{sec:self_supervised} presents a language model adaptation process based on maximum-likelihood and maximum-aposteriori self-training, 
while Section \ref{sec:asr_channel_approach}
introduces adaptation that utilizes ASR channel estimates. Section \ref{sec:experiments} describes experiments on conversational speech.

\section{Language Model Adaptation}
\label{sec:problem_formulation}


We are given a trained speech recognizer, topic-based language model and a large collection
of audio utterances (many conversations) for a new domain, i.e. a change of topics, but not manually transcribed text needed for language model training.
Our goal is to adapt the language model by learning new topic distributions
using the available audio. We consider self-training that adapts the topic distributions based on
automatically transcribed audio.

A {\em conversation} ${\cal C}$ is composed of $N$ speech utterances are represented by
$N$ lattices (confusion networks) -- annotated with words and posterior probabilities -- produced by the speech recognizer.
\footnote{We assume conversations but our methods can be applied to non-conversational genres.} Each confusion network consists of a 
sequence of {\em bins}, where each bin is a set of words hypothesized by the recognizer at a particular time.
The $i$-th bin is denoted by $b_i$ and contains words $\{w_{i,j}\}_{j=1}^{|\cal{V}|}$, where ${\cal V}$ is the vocabulary.
\footnote{Under {\em specific} contexts, not all vocabulary items are likely to be truly spoken in a bin. Although we use summations over all $w 
\in {\cal V}$, we practically use only words which are acoustically confusable and consistent with the lexical context of the bin.} When obvious from context, we omit 
subscript $j$.
The most likely word in bin $b_i$ is denoted by $\widehat{w}_i$. Let $M$ be the total number of bins in the $N$ confusion networks in a single conversation.

We use unigram topic-based language models (multinomial distributions over $\cal{V}$), 
which capture word frequencies under each topic. Such models have been used in a variety of ways,
such as in PLSA \cite{Hofmann01} and LDA \cite{BNJ03}, and under different training scenarios.
Topic-based models provide a global context beyond the local word 
history in modeling a word's probability, and have been found especially useful in language model adaptation
\cite{TS2005,WS2007,HG06}.
Each topic $t \in \{1, \ldots, T\}$ has a multinomial distribution
denoted by $q(w|t), w \in {\cal V}$. These topic-word distributions are learned from conversations labeled with topics,
such as those in the Fisher speech corpus.\footnote{Alternative approaches estimate topics from unlabeled data, but we use labeled
data for evaluation purposes.}

Adapting the topic-based language model means learning a set of conversation specific mixture weights $\boldsymbol{\lambda}^{\cal C} = 
(\lambda_1^{\cal C}, \ldots, 
\lambda_T^{\cal C})$, where $\lambda_t^{\cal C}$ indicates the likelihood of seeing topic $t$ in conversation ${\cal C}$.\footnote{Unless 
required we drop the superscript ${\cal C}$.} While topic compositions remain fixed, the topics selected change with each conversation.\footnote
{This assumption can be relaxed to learn $q(w|t)$ as well.} These mixture weights form the true distribution of a word:
\begin{equation}
q(w) = \sum_{t=1}^T \lambda_t q(w|t).
\label{eq:prior}
\end{equation}

\section{Learning $\boldsymbol{\lambda}$ from ASR Output}
\label{sec:self_supervised}


We begin by following previous approaches to self-training, in which model parameters are re-estimated based on ASR output. We consider
self-training based on 1-best and lattice maximum-likelihood estimation (MLE) as well as maximum-aposteriori (MAP) training. In the next section, we  modify these approaches to incorporate our confusion estimates.

For estimating the topic mixtures $\boldsymbol{\lambda}$ using self-training on the 1-best ASR output, i.e.
the 1-best path in the confusion network $\mathbf{\widehat{w}} = \{\widehat{w}_i\}_{i=1}^M$, we write
the log-likelihood of the observations $\mathbf{\widehat{w}}$:
\begin{equation}
{\cal L}^{(1)} = \sum_{i=1}^M \log q(\widehat{w}_i),
\label{eq:log_like_l1}
\end{equation}
where $q$ is a mixture of topic models (\ref{eq:prior}),
for mixture weights $\boldsymbol{\lambda}$. We expect topic-based distributions will better estimate the true word 
distribution than the empirical estimate
$
\hat{q}(w) = \frac{1}{M} \sum_{i=1}^M \boldsymbol{1}(\widehat{w}_i = w),
$ 
as the latter is biased due to ASR errors.
Maximum-likelihood estimation of $\boldsymbol{\lambda}$ in (\ref{eq:log_like_l1}) is given by:
\begin{eqnarray}
\boldsymbol{\lambda}^* & = & \arg\max_{\boldsymbol{\lambda}}  \sum_{i=1}^M \log q(\widehat{w}_i)\nonumber\\
& = & \arg\max_{\boldsymbol{\lambda}}  \sum_{i=1}^M \log \left(\sum_{t} \lambda_t q(\widehat{w}_i|t)\right)
\label{eq:argmax1}
\end{eqnarray}

Using the EM algorithm, $Q$ is the expected log-likelihood of the ``complete'' data:
\begin{equation}
Q(\boldsymbol{\lambda}^{(j)}, \boldsymbol{\lambda}) = \sum_{i=1}^M \sum_t r^{(j)}(t | \widehat{w}_i) \log(\lambda_t q(\widehat{w}_i | t)),
\label{eq:q_func1}
\end{equation}
where $r^{(j)}(t | \widehat{w}_i)$ is the posterior distribution of the topic in the $i$-th bin, given the 1-best word $\widehat{w}_i$; this is computed in the E-step of the $j$-th iteration:
\begin{equation}
r^{(j)}(t | \widehat{w}_i) = \frac{\lambda_t^{(j)} q(\widehat{w}_i | t)}{\sum_{t^\prime} \lambda_{t^\prime}^{(j)} q(\widehat{w}_i | t^\prime)}.
\label{eq:posterior1}
\end{equation}
In the M-step of the $j+1$ iteration, the new estimate of the prior is computed by maximizing (\ref{eq:q_func1}), i.e.,
\begin{equation}
\lambda_t^{(j+1)} = \frac{\sum_{i=1}^M r^{(j)}(t | \widehat{w}_i)}{\sum_{t^\prime} \sum_{i^\prime=1}^M r^{(j)}(t^\prime | \widehat{w}_{i^\prime})}.
\label{eq:lambda_update_1}
\end{equation}

\subsection{Learning $\boldsymbol{\lambda}$ from Expected Counts}
\label{sec:self_supervised_expected_counts}
Following \newcite{nsm09} we next consider using the entire bin in self-training by
maximizing the {\em expected} log-likelihood of the ASR output:
\begin{equation}
{\cal L}^{\prime (1)} = E\left[\sum_{i=1}^M \log q(W_i)\right],
\label{eq:expectation}
\end{equation}
where $W_i$ is a random variable which takes the value $w \in b_i$ with probability equal to the confidence (posterior probability)
$s_i(w)$ of the recognizer.
The maximum-likelihood estimation problem becomes:
\begin{eqnarray}
\boldsymbol{\lambda}^*\!&\!\!\!\!=\!\!\!\!& \arg\max_{\boldsymbol{\lambda}} {\cal L}^{\prime (1)}\\
\!\!\!\!&\!\!\!\!=\!\!\!\!&\arg\max_{\boldsymbol{\lambda}}  \sum_{i=1}^M \sum_{w \in b_i} s_i(w) \log \left(\sum_{t} \lambda_t q(w|t)\right)\nonumber\\
&\!\!\!\!=\!\!\!\!&\arg\max_{\boldsymbol{\lambda}}  \sum_{w \in {\cal V}} \tf(w) \log \left(\sum_{t} \lambda_t q(w|t)\right)\nonumber
\label{eq:argmax2}
\end{eqnarray}
where $\tf(w) = \sum_i s_i(w)$ denotes the {\em expected count} of word $w$ in the conversation,
given by the sum of the posteriors of $w$
in all the confusion network bins of the conversation \cite{KDCJK11} (note that for text documents, it is equivalent to term-frequency).
We again use the EM algorithm, with objective function:
\begin{equation}
Q(\boldsymbol{\lambda}^{(j)}, \boldsymbol{\lambda}) = \sum_{w \in {\cal V}} \tf(w) \sum_t r^{(j)}(t | w) \log(\lambda_t q(w | t)),
\label{eq:q_func2}
\end{equation}
where $r^{(j)}(t | w)$ is the posterior distribution of the topic given word $w$ computed using (\ref{eq:posterior1})
(but with $w$ instead of $\widehat{w}_i$).
In the M-step of the $j+1$ iteration, the new estimate of the prior is computed by maximizing (\ref{eq:q_func2}), i.e.,
\begin{equation}
\lambda_t^{(j+1)} = \frac{\sum_{w \in {\cal V}} \tf(w) r^{(j)}(t | w)}{\sum_{t^\prime} \sum_{w \in {\cal V}} \tf(w) r^{(j)}(t^\prime | w)}.
\label{eq:lambda_update_2}
\end{equation}

\subsection{Maximum-Aposteriori Estimation of $\boldsymbol{\lambda}$}
\label{sec:prior}
In addition to a maximum likelihood estimate, we consider a maximum-aposteriori (MAP) estimate
by placing a Dirichlet prior over $\boldsymbol{\lambda}$ \cite{BR2003}:
\begin{equation}
\boldsymbol{\lambda}^* = \arg\max_{\boldsymbol{\lambda}} {\cal L} + \beta \log(\text{Dir}(\boldsymbol{\lambda}; \alpha))
\label{eq:map_argmax1}
\end{equation}
where $\text{Dir}(\boldsymbol{\lambda}; \alpha)$ is the pdf of the Dirichlet distribution with parameter $\alpha$:
$
\text{Dir}(\boldsymbol{\lambda}; \alpha) = \frac{1}{B(\alpha)} \prod_t \lambda_t^{\alpha-1}.
$ 

This introduces an extra component in the optimization. It is easy to prove that the update equation for $\lambda_t^{(j+1)}$ becomes:
\begin{equation}
\lambda_t^{(j+1)}\!=\!\left[\frac{\sum_{i=1}^M r^{(j)}(t | \widehat{w}_i) + \beta(\alpha-1)}{\sum_{t^\prime} \sum_{i=1}^M r^{(j)}(t^\prime | \widehat{w}_i) + T\beta(\alpha-1)}\right]_+.
\label{eq:lambda_map_1}
\end{equation}
for the case where only the ASR 1-best is used, and
\begin{equation}
\lambda_t^{(j+1)}\!=\!\left[\frac{\sum_{w \in {\cal V}} \tf(w) r^{(j)}(t | w) + \beta(\alpha-1)}{\sum_{t^\prime} \sum_{w \in {\cal V}} \tf(w) r^{(j)}(t^\prime | w) + T \beta(\alpha-1)}\right]_+.
\label{eq:lambda_map_2}
\end{equation}
for the case where expected counts are used.
Note that the $[x]_+$ notation stands for $\max\{0,x\}$.

\section{Confusion Estimation in ASR Output}
\label{sec:asr_channel_approach}
\newcommand{\tikzsize}{.8}
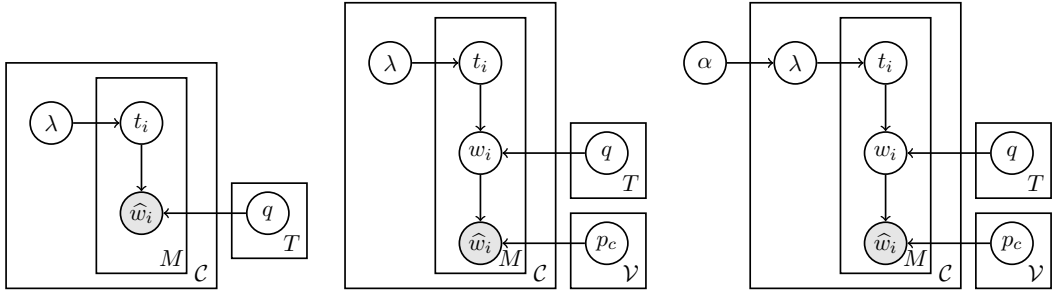
\begin{figure*}[tb]
 \begin{center}
  \scalebox{\tikzsize}{
    \tikzstyle{var}=[circle,draw=black!100,fill=none,thick,minimum
size=7mm,inner sep=0mm]
    \tikzstyle{hidden}=[draw]
    \tikzstyle{obs}=[circle,draw=black!100,fill=none,thick,minimum
size=7mm,fill=gray!20,inner sep=0mm]
    \begin{tikzpicture}[style=thick]
      \draw (2.25,1.25) rectangle (3.75,4.5);
      \draw (0.75,1) rectangle (4.25,4.75);

            \draw (4.5,1.5) rectangle (5.75,2.75);

      \node (q) at (5.1,2.25) [var] {$q$};
      \node (hatw) at (3,2.25) [obs] {$\widehat{w}_i$};      
      \node (t) at (3.0,3.75) [var] {$t_i$};
      \node (lambda) at (1.5,3.75) [var] {$\lambda$};

      \node (M) at (3.5,1.5) {$M$};
      \node (C) at (4,1.25) {$\cal{C}$};
      \node (T) at (5.5,1.75) {$T$};

\draw [->] (lambda) -- (t);
\draw [->] (t) -- (hatw);
\draw [->] (q) -- (hatw);    \end{tikzpicture}
  }
  \scalebox{\tikzsize}{
    \tikzstyle{var}=[circle,draw=black!100,fill=none,thick,minimum
size=7mm,inner sep=0mm]
    \tikzstyle{hidden}=[draw]
    \tikzstyle{obs}=[circle,draw=black!100,fill=none,thick,minimum
size=7mm,fill=gray!20,inner sep=0mm]
    \begin{tikzpicture}[style=thick]
      \draw (2.25,.25) rectangle (3.75,4.5);
      \draw (0.75,0) rectangle (4.25,4.75);
      \draw (4.5,0) rectangle (5.75,1.25);
            \draw (4.5,1.5) rectangle (5.75,2.75);

      \node (hatw) at (3.0,0.75) [obs] {$\widehat{w}_i$};
      \node (psubc) at (5.1,0.75) [var] {$p_c$};
      \node (q) at (5.1,2.25) [var] {$q$};
      \node (w) at (3,2.25) [var] {$w_i$};      
      \node (t) at (3.0,3.75) [var] {$t_i$};
      \node (lambda) at (1.5,3.75) [var] {$\lambda$};

      \node (M) at (3.5,.5) {$M$};
      \node (C) at (4,.25) {$\cal{C}$};
      \node (vocab) at (5.5,.25) {$\cal{V}$};
      \node (T) at (5.5,1.75) {$T$};

\draw [->] (lambda) -- (t);
\draw [->] (t) -- (w);
\draw [->] (w) -- (hatw);
\draw [->] (psubc) -- (hatw);
\draw [->] (q) -- (w);    \end{tikzpicture}
  }
  \scalebox{\tikzsize}{
    \tikzstyle{var}=[circle,draw=black!100,fill=none,thick,minimum
size=7mm,inner sep=0mm]
    \tikzstyle{hidden}=[draw]
    \tikzstyle{obs}=[circle,draw=black!100,fill=none,thick,minimum
size=7mm,fill=gray!20,inner sep=0mm]
    \begin{tikzpicture}[style=thick]
      \draw (2.25,.25) rectangle (3.75,4.5);
      \draw (0.75,0) rectangle (4.25,4.75);
      \draw (4.5,0) rectangle (5.75,1.25);
            \draw (4.5,1.5) rectangle (5.75,2.75);

      \node (hatw) at (3.0,0.75) [obs] {$\widehat{w}_i$};
      \node (psubc) at (5.1,0.75) [var] {$p_c$};
      \node (q) at (5.1,2.25) [var] {$q$};
      \node (w) at (3,2.25) [var] {$w_i$};      
      \node (t) at (3.0,3.75) [var] {$t_i$};
      \node (lambda) at (1.5,3.75) [var] {$\lambda$};
      \node (alpha) at (0,3.75) [var] {$\alpha$};

      \node (M) at (3.5,.5) {$M$};
      \node (C) at (4,.25) {$\cal{C}$};
      \node (vocab) at (5.5,.25) {$\cal{V}$};
      \node (T) at (5.5,1.75) {$T$};

\draw [->] (alpha) -- (lambda);
\draw [->] (lambda) -- (t);
\draw [->] (t) -- (w);
\draw [->] (w) -- (hatw);
\draw [->] (psubc) -- (hatw);
\draw [->] (q) -- (w);    \end{tikzpicture}
  }
 \end{center}
\caption{The graphical model representations for maximum-likelihood 1-best self-training (left), self-training with the ASR channel confusion 
model (middle), MAP training with the confusion model (right).
In all models, each word is conditionally independent given the selected topic,
and topic proportions $\boldsymbol{\lambda}$ are conversation specific. 
For the middle and right models, each observed top word in the $i$-th confusion network bin ($\widehat{w}_i$) is generated by first sampling
a topic $t_i$ according to $\boldsymbol{\lambda}$, then sampling the (true) word $w_i$ according to the topic $t_i$, and finally sampling 
the observed ASR word $\widehat{w}_i$ based on the ASR channel confusion model ($p_c(v|w)$). For expected count models, $\widehat{w}_i$ becomes
a \emph{distribution} over words in the confusion network bin $b_i$.}
\label{fig:graphical_model}
\end{figure*}

Self-training on ASR output can mislead the language model through confusions in the ASR channel. By modeling these confusions
we can guide self-training and recover from recognition errors.

The ASR channel confusion model is represented by 
a conditional probability distribution $p_c(v|w), v,w \in {\cal V}$, which denotes the probability that the most likely word in the output 
of the recognizer (i.e., the ``1-best'' word) is $v$, given that the true (reference) word spoken is $w$.
Of course, this conditional distribution is just an approximation as many other phenomena -- coarticulation, non-stationarity of
speech, channel variations, lexical choice in the context, etc. -- cause this probability to vary. We assume that $p_c(v|w)$
is an ``average'' of the conditional probabilities under various conditions. 

We use the following simple procedure for estimating the ASR channel, similar to that of \cite{XKK09} for computing cohort sets:
\squishlist
\item Create confusion networks \cite{mbs99} with the available audio.
\item Count $c(w,v)$, the number of times words $w,v$ appear in the same bin.
\item The conditional ASR probability is computed as $p_c(v|w) = c(v,w)/\sum_{v'} c(v',w)$.
\item Prune words whose posterior probabilities are lower than 5\% of the max probability in a bin.
\item Keep only the top 10 words in each bin.
\squishend

The last two items above were implemented as a way of reducing the search space and the complexity of the task. We did not observe significant changes in the results when we relaxed these two constraints.

\subsection{Learning $\boldsymbol{\lambda}$ with ASR Confusion Estimates}
We now derive a maximum-likelihood estimate of $\boldsymbol{\lambda}$ based on the 1-best ASR output but relies on the ASR channel confusion model $p_c$.
The log-likelihood of the observations 
(most likely path in the confusion network) is:
\begin{equation}
{\cal L}^{(2)} = \sum_{i=1}^M \log p(\widehat{w}_i),
\end{equation}
where $p$ is the induced distribution on the observations under the confusion model $p_c(v|w)$ and the estimated distribution $q$ of
(\ref{eq:prior}):
\begin{equation}
p(\widehat{w}_i) = \sum_{w \in \cal{V}} q(w) p_c(\widehat{w}_i|w).
\label{eq:marginal1}
\end{equation}
Recall that while we sum over $\cal{V}$, in practice the summation is limited to only likely words. One could argue that the ASR channel 
confusion distribution $p_c(\widehat{w}_i|w)$ should discount unlikely confusions.
However, since $p_c$ is not bin specific, unlikely words in a specific bin
could receive non-negligible probability from $p_c$ if they are likely in general. This makes the
truncated summation over $\cal{V}$ problematic.

One solution would be to reformulate $p_c(\widehat{w}_i|w)$ so that it becomes a conditional distribution given 
the left (or even right) lexical context of the bin. But this formulation adds complexity and
suffers from the usual sparsity issues. The solution we follow here imposes the constraint that only the words already existing in 
$b_i$ are allowed to be candidate words giving rise to $\widehat{w}_i$. 
This uses the ``pre-filtered'' set of words in $b_i$ to condition on context (acoustic and lexical), without having to model such context explicitly. 
We anticipate that this conditioning on the words of $b_i$ leads to more accurate inference.
The likelihood objective then becomes:
\begin{equation}
{\cal L}^{\prime (2)} = \sum_{i=1}^M \log p_i(\widehat{w}_i),
\end{equation}
with $p_i$ defined as:
\begin{equation}
p_i(\widehat{w}_i) = \sum_{w \in b_i} \frac{q(w)}{\sum_{w^\prime \in b_i} q(w^\prime)} p_c(\widehat{w}_i|w),
\label{eq:marginal2}
\end{equation}
i.e., the induced probability conditioned on bin $b_i$. 
Note that although we estimate a conversation level distribution $q(w)$, it has to be normalized in each bin by dividing by
$\sum_{w \in b_i} q(w)$, in order to condition only on the words in the bin.
The maximum-likelihood estimation for $\boldsymbol{\lambda}$ becomes:
\begin{eqnarray}
\lefteqn{\mspace{-170mu}\boldsymbol{\lambda}^* = \arg\max_{\boldsymbol{\lambda}} {\cal L}^{\prime (2)}}\nonumber
\lefteqn{=\arg\max_{\boldsymbol{\lambda}}  \sum_{i=1}^M \log p_i(\widehat{w}_i)}\nonumber\\
\lefteqn{\mspace{-170mu}=\arg\max_{\boldsymbol{\lambda}}  \sum_{i=1}^M \log \left(\sum_{w \in b_i} \frac{\sum_{t} \lambda_t q(w|t) p_c(\widehat{w}_i | w)}{\sum_{w^\prime \in b_i} \sum_{t^\prime} \lambda_{t^\prime} q(w^\prime|t^\prime)}\right)}\nonumber\\
\label{eq:argmax}
\end{eqnarray}
Note that the appearance of $\lambda_t$ in the denominator makes the maximization harder.

As before, we rely on an EM-procedure to maximize this objective. Let us assume that at the $j$-th iteration of EM we have an estimate of the prior distribution $q$, denoted by $q^{(j)}$. This induces an observation probability in bin $i$ based on equation (\ref{eq:marginal2}), as well as a posterior probability $r_i^{(j)}(w | \widehat{w}_i)$ that a word $w \in b_i$ is the reference word, given $\widehat{w}_i$. The goal is to come up with a new estimate of the prior distribution $q^{(j+1)}$ that increases the value of the log-likelihood. 
\remove{
We can re-write the objective function:
\begin{eqnarray}
{\cal L}^{\prime (2)}
&\!\!\!\!=\!\!\!\!& \sum_{i=1}^M \sum_{w \in b_i} r_i^{(j)}(w | \widehat{w}_i) \log p_i(\widehat{w}_i)\nonumber\\
&\!\!\!\!=\!\!\!\!& \sum_{i=1}^M \sum_{w \in b_i} r_i^{(j)}(w | \widehat{w}_i) \log \left(\frac{p_i^{\prime}(\widehat{w}_i) r_i^{\prime}(w | \widehat{w}_i)}{r_i^{\prime}(w | \widehat{w}_i)}\right)\nonumber\\
&\!\!\!\!=\!\!\!\!& \sum_{i=1}^M \sum_{w \in b_i} r_i^{(j)}(w | \widehat{w}_i) \log (p_c(\widehat{w}_i | w) q_i^{\prime}(w))\nonumber\\
&&- \sum_{i=1}^M \sum_{w \in b_i} r_i^{(j)}(w | \widehat{w}_i) \log(r_i^{\prime}(w | \widehat{w}_i))
\label{eq:em1}
\end{eqnarray}
We denote (\ref{eq:em1}) by ${\cal L}^{\prime (2)}(q^{(j)},q^{\prime})$, and our goal is to come up with a value for $q^{\prime}$ (that will become $q^{(j+1)}$) that maximizes ${\cal L}^{\prime (2)}(q^{(j)},q^{\prime})$, or, equivalently, maximizes the difference ${\cal L}^{\prime (2)}(q^{(j)},q^{\prime}) - {\cal L}^{\prime (2)}(q^{(j)}, q^{(j)})$. 
}
If we define:
\begin{eqnarray}
\lefteqn{Q(q^{(j)},q^{\prime}) =}\nonumber\\
\!\!\!\!&=&\!\!\!\!\sum_{i=1}^M \sum_{w \in b_i} r_i^{(j)}(w | \widehat{w}_i) \log (p_c(\widehat{w}_i | w) q_i^{\prime}(w))\nonumber\\
\!\!\!\!&=&\!\!\!\!\sum_{i=1}^M \sum_{w \in b_i} r_i^{(j)}(w | \widehat{w}_i) \log \left(\frac{p_c(\widehat{w}_i | w) q^{\prime}(w)}{\sum_{w^\prime \in b_i} q^{\prime}(w^\prime)}\right)\nonumber\\
\mspace{-40mu}
\label{eq:q1}
\end{eqnarray}
then we have:
\begin{eqnarray}
\lefteqn{{\cal L}^{\prime (2)}(q^{(j)},q^{\prime}) - {\cal L}^{\prime (2)}(q^{(j)}, q^{(j)})}\nonumber\\
& = & Q(q^{(j)},q^{\prime})-Q(q^{(j)},q^{(j)})\nonumber\\
&&+\sum_{i=1}^M D(r_i^{(j)}(\cdot | \widehat{w}_i) \| r_i^{\prime}(\cdot | \widehat{w}_i))\nonumber\\
& \geq & Q(q^{(j)},q^{\prime})-Q(q^{(j)},q^{(j)}) ~,
\label{eq:q_diff}
\end{eqnarray}
where (\ref{eq:q_diff}) holds because $D(r_i^{(j)}(\cdot | \widehat{w}_i) \| r_i^{\prime}(\cdot | \widehat{w}_i)) \geq 0$ \cite{CT96}. Thus, we just need to find the value of $q^\prime$ that maximizes $Q(q^{(j)}, q^{\prime})$, as this will guarantee that ${\cal L}^{\prime (2)}(q^{(j)},q^{\prime}) \geq {\cal L}^{\prime (2)}(q^{(j)}, q^{(j)})$.
The distribution $q^\prime$ can be written:
\begin{equation}
q^\prime(w) = \sum_t \lambda_t q(w|t).
\end{equation}
Thus, $Q$, as a function of $\boldsymbol{\lambda}$, can be written as:
\begin{eqnarray}
\lefteqn{Q(q^{(j)},\boldsymbol{\lambda}) =}\nonumber\\
\mspace{-40mu}&&\mspace{-40mu}\sum_{i=1}^M \sum_{w \in b_i} r_i^{(j)}(w | \widehat{w}_i) \log \left(\frac{p_c(\widehat{w}_i | w) \sum_t \lambda_t q(w|t)}{\sum_{w^\prime \in b_i} \sum_t \lambda_t q(w^\prime|t)}\right).\nonumber\\
\label{eq:q2}
\end{eqnarray}
Equation (\ref{eq:q2}) cannot be easily maximized with respect to $\boldsymbol{\lambda}$ by simple differentiation, because the elements of $\boldsymbol{\lambda}$ appear in the denominator, making them coupled. Instead, we will derive and maximize a lower bound for the $Q$-difference (\ref{eq:q_diff}).
\remove
{
First, we prove the following lemma.
\begin{lemma}
For any set of positive numbers $\{a(1), \ldots, a(T)\}$, and $\{b(1), \ldots, b(T)\}$, we have
\begin{equation}
\log\left(\frac{\sum_t a(t) b(t)}{\sum_t a(t)}\right) \geq \frac{\sum_t \log(b(t)) a(t)}{\sum_t a(t)}.
\end{equation}
\label{lemma1}
\end{lemma}
\begin{proof}
Jensen's inequality states that, for a concave function $f$ and a distribution $p$,
\begin{equation}
f(E[X]) \geq E[f(X)].
\end{equation}
Setting $p(t) = a(t)/\sum_t^\prime a(t^\prime)$, and because $\log(\cdot)$ is a concave function, we obtain
\begin{equation}
\log\left(\sum_t p(t) b(t)\right) \geq \sum_t p(t) \log(b(t)),
\end{equation}
which is the desired relationship.
\end{proof}
}

For the rest of the derivation we assume that $\lambda_t = \frac{e^{\mu_t}}{\sum_{t^\prime} e^{\mu_{t^\prime}}}$, where $\mu_t \in {\cal R}$.
We can thus express $Q$ as a function of $\boldsymbol{\mu} = (\mu_1, \ldots, \mu_T)$ as follows:
\begin{eqnarray}
\lefteqn{Q(q^{(j)},\boldsymbol{\mu}) =}\nonumber\\
&&\mspace{-35mu}\sum_{i=1}^M \sum_{w \in b_i} r_i^{(j)}(w | \widehat{w}_i) \log \left(\frac{p_c(\widehat{w}_i | w) \sum_t \frac{e^{\mu_t}}{\sum_{t^\prime} e^{\mu_{t^\prime}}} q(w|t)}{\sum_{w^\prime \in b_i} \sum_t \frac{e^{\mu_t}}{\sum_{t^\prime} e^{\mu_{t^\prime}}} q(w^\prime|t)}\right)\mspace{-200mu}\nonumber\\
&\mspace{-20mu}=&\mspace{-15mu}\sum_{i=1}^M \sum_{w \in b_i} r_i^{(j)}(w | \widehat{w}_i) \log \left(\frac{p_c(\widehat{w}_i | w) \sum_t e^{\mu_t} q(w|t)}{\sum_{w^\prime \in b_i} \sum_t e^{\mu_t} q(w^\prime|t)}\right)\nonumber\\
\label{eq:q3}
\end{eqnarray}
Interestingly, the fact that the sum $\sum_t e^{\mu_t}$ appears in both numerator and denominator in the above expression allows us to discard it.

At iteration $j+1$, the goal is to come up with an update $\boldsymbol{\delta}$, such that $\boldsymbol{\mu}^{(j)}+\boldsymbol{\delta}$ results in a higher value for $Q$; i.e., we require:
\begin{equation}
Q(q^{(j)}, \boldsymbol{\mu}^{(j)}+\boldsymbol{\delta}) \geq Q(q^{(j)}, \boldsymbol{\mu}^{(j)}),
\end{equation}
where $\boldsymbol{\mu}^{(j)}$ is the weight vector that resulted after optimizing $Q$ in the $j$-th iteration. Obviously,
\begin{equation}
q^{(j)}(w) = \sum_t \frac{e^{\mu_t^{(j)}} q(w|t)}{\sum_{t^\prime} e^{\mu_{t^\prime}^{(j)}}}.
\end{equation}
Let us consider the difference:
\begin{eqnarray}
\lefteqn{Q(q^{(j)}, \boldsymbol{\mu}^{(j)}+\boldsymbol{\delta}) - Q(q^{(j)}, \boldsymbol{\mu}^{(j)})}
\label{eq:q_diff_new}\\
& = & \sum_{i=1}^M \sum_{w \in b_i} r_i^{(j)}(w | \widehat{w}_i) \times\nonumber\\
&&\Bigg[-\log \left(\frac{\sum_t e^{\mu_t^{(j)}+\delta_t} \sum_{w^\prime \in b_i} q(w^\prime|t)}{\sum_t e^{\mu_t^{(j)}} \sum_{w^\prime \in b_i} q(w^\prime|t)}\right)\nonumber\\
&&+ \log\left(\frac{\sum_t e^{\mu_t^{(j)}+\delta_t} q(w|t)}{\sum_t e^{\mu_t^{(j)}} q(w|t)}\right)\Bigg] ~.
\label{eq:q_diff_new2}
\end{eqnarray}
We use the well-known inequality $\log(x) \leq x-1 \Rightarrow -\log(x) \geq 1-x$ and obtain:
\begin{eqnarray}
\lefteqn{-\log \left(\frac{\sum_t e^{\mu_t^{(j)}+\delta_t} \sum_{w^\prime \in b_i} q(w^\prime|t)}{\sum_t e^{\mu_t^{(j)}} \sum_{w^\prime \in b_i} q(w^\prime|t)}\right) \geq}\nonumber\\
&& 1 - \frac{\sum_t e^{\mu_t^{(j)}+\delta_t} \sum_{w^\prime \in b_i} q(w^\prime|t)}{\sum_t e^{\mu_t^{(j)}} \sum_{w^\prime \in b_i} q(w^\prime|t)}.
\label{eq:ineq1}
\end{eqnarray}
\remove{
Next, we set
$
a(t) \triangleq e^{\mu_t^{(j)}} q(w|t),\ \ b(t) \triangleq e^{\delta_t},
$ 
}
Next, we apply Jensen's inequality to obtain:
\begin{equation}
\log\left(\frac{\sum_t e^{\mu_t^{(j)}+\delta_t} q(w|t)}{\sum_t e^{\mu_t^{(j)}} q(w|t)}\right) \geq \frac{\sum_t e^{\mu_t^{(j)}} q(w|t) \delta_t}{\sum_t e^{\mu_t^{(j)}} q(w|t)}.
\label{eq:ineq2}
\end{equation}
By combining (\ref{eq:q_diff_new2}), (\ref{eq:ineq1}) and (\ref{eq:ineq2}), we obtain a lower bound on the $Q$-difference of (\ref{eq:q_diff_new}):
\begin{eqnarray}
\lefteqn{Q(q^{(j)}, \boldsymbol{\mu}^{(j)}+\boldsymbol{\delta}) - Q(q^{(j)}, \boldsymbol{\mu}^{(j)}) \geq}\nonumber\\
\mspace{-40mu}&&\mspace{-35mu}\sum_{i=1}^M \sum_{w \in b_i} r_i^{(j)}(w | \widehat{w}_i)\Bigg[1 + \frac{\sum_t e^{\mu_t^{(j)}} q(w|t) \delta_t}{\sum_t e^{\mu_t^{(j)}} q(w|t)} \nonumber\\
\mspace{-40mu}&&\mspace{-35mu} -\frac{\sum_t e^{\mu_t^{(j)}+\delta_t} \sum_{w^\prime \in b_i} q(w^\prime|t)}{\sum_t e^{\mu_t^{(j)}} \sum_{w^\prime \in b_i} q(w^\prime|t)}\Bigg] ~.
\label{eq:q_diff_lb}
\end{eqnarray}
It now suffices to find a $\boldsymbol{\delta}$ that maximizes the lower bound of (\ref{eq:q_diff_lb}), as this will guarantee that the $Q$-difference will be greater than zero. Note that the lower bound in (\ref{eq:q_diff_lb}) is a {\em concave} function of $\delta_t$, and it thus has a global maximum that we will find using differentiation. Let us set $g(\boldsymbol{\delta})$ equal to the right-hand-side of (\ref{eq:q_diff_lb}). Then,
\begin{eqnarray}
\lefteqn{\frac{\partial g(\boldsymbol{\delta})}{\partial \delta_t} = \sum_{i=1}^M \sum_{w \in b_i} r_i^{(j)}(w | \widehat{w}_i) \times}\nonumber\\ 
&&\mspace{-35mu}\left[-\frac{e^{\mu_{t}^{(j)}+\delta_t} \sum_{w^\prime \in b_i} q(w^\prime|t)}{\sum_{t^\prime} e^{\mu_{t^\prime}^{(j)}} \sum_{w^\prime \in b_i} q(w^\prime | t^\prime))} + \frac{e^{\mu_{t}^{(j)}} q(w|t)}{\sum_{t^\prime} e^{\mu_{t^\prime}^{(j)}} q(w|t^\prime) }\right].\nonumber\\
\label{eq:partial1}
\end{eqnarray}
By setting (\ref{eq:partial1}) equal to 0 and solving for $\delta_t$, we obtain the update for $\delta_t$ (or, equivalently, $e^{\mu_{t}^{(j)}+\delta_t}$):
\begin{eqnarray}
\lefteqn{\mspace{-40mu}e^{\mu_{t}^{(j)}+\delta_t} = \Bigg(\sum_{i=1}^M \sum_{w \in b_i} \frac{r_i^{(j)}(w | \widehat{w}_i) e^{\mu_{t}^{(j)}} q(w|t)}{\sum_{t^\prime} e^{\mu_{t^\prime}^{(j)}}q(w|t^\prime)}\Bigg)}\nonumber\\
&&\mspace{-40mu}\times\Bigg(\sum_{i=1}^M \frac{\sum_{w \in b_i} q(w|t)}{\sum_{t^\prime} e^{\mu_{t^\prime}^{(j)}} \sum_{w \in b_i} q(w|t^\prime)}\Bigg)^{\!\!-1}
\label{eq:update1}
\end{eqnarray}

\subsection{Learning $\boldsymbol{\lambda}$ from Expected Counts}
\label{sec:asr_channel_approach_expected_counts}
As before, we now consider 
maximizing the expected log-likelihood of the ASR output:
\begin{equation}
{\cal L}^{\prime\prime (2)} = E\left[\sum_{i=1}^M \log p(W_i)\right],
\label{eq:expectation_part2}
\end{equation}
where $W_i$ takes the value $w \in b_i$ with probability equal to the confidence (posterior probability)
$s_i(w)$ of the recognizer.
The modified maximum-likelihood estimation problem now becomes:
\begin{eqnarray}
\lefteqn{\mspace{-0mu}\boldsymbol{\lambda}^* = \arg\max_{\boldsymbol{\lambda}} {\cal L}^{\prime\prime (2)}}\nonumber\\
\lefteqn{\mspace{-0mu}=\arg\max_{\boldsymbol{\lambda}}  \sum_{i=1}^M \sum_{w \in b_i} s_i(w) \log p_i(w)}\nonumber\\
\lefteqn{\mspace{-0mu}=\arg\max_{\boldsymbol{\lambda}}  \sum_{i=1}^M \sum_{w \in b_i} s_i(w) \times} \nonumber\\
&&\log \left(\sum_{w^\prime \in b_i} \frac{\sum_{t} \lambda_t q(w^\prime|t) p_c(w | w^\prime)}{\sum_{w^{\prime\prime} \in b_i} \sum_{t^\prime} \lambda_{t^\prime} q(w^{\prime\prime}|t^\prime)}\right) ~.\nonumber\\
\label{eq:argmax_part2}
\end{eqnarray}
By following a procedure similar to the one described earlier in this section, we come up with the lower bound on the $Q$-difference:
\begin{eqnarray}
\lefteqn{Q(q^{(j)}, \boldsymbol{\mu}^{(j)}+\boldsymbol{\delta}) - Q(q^{(j)}, \boldsymbol{\mu}^{(j)}) \geq}\nonumber\\
\mspace{-40mu}&&\mspace{-35mu}\sum_{i=1}^M \sum_{w,w^\prime \in b_i} r_i^{(j)}(w | w^\prime) s_i(w^\prime) \Bigg[1+\frac{\sum_t e^{\mu_t^{(j)}} q(w|t) \delta_t}{\sum_t e^{\mu_t^{(j)}} q(w|t)}\nonumber\\
\mspace{-40mu}&&\mspace{-35mu}- \frac{\sum_t e^{\mu_t^{(j)}+\delta_t} \sum_{w^\prime \in b_i} q(w^\prime|t)}{\sum_t e^{\mu_t^{(j)}} \sum_{w^\prime \in b_i} q(w^\prime|t)}\Bigg]
\label{eq:q_diff_lb_ec}
\end{eqnarray}
whose optimization results in the following update equation for $\delta_t$:
\begin{eqnarray}
\lefteqn{\mspace{-0mu}e^{\mu_{t}^{(j)}+\delta_t} =}\nonumber\\
&&\Bigg(\sum_{i=1}^M \sum_{w,w^\prime \in b_i} \frac{r_i^{(j)}(w | w^\prime) s_i(w^\prime)  e^{\mu_{t}^{(j)}} q(w|t)}{\sum_{t^\prime} e^{\mu_{t^\prime}^{(j)}}q(w|t^\prime)}\Bigg)\nonumber\\
&&\mspace{-0mu}\times\Bigg(\sum_{i=1}^M \frac{\sum_{w \in b_i} q(w|t)}{\sum_{t^\prime} e^{\mu_{t^\prime}^{(j)}} \sum_{w \in b_i} q(w|t^\prime)}\Bigg)^{\!\!-1}~.
\label{eq:update1_part2}
\end{eqnarray}

\subsection{Maximum-Aposteriori Estimation of $\boldsymbol{\lambda}$}
\label{sec:map_version}
Finally, we consider a MAP estimation of $\boldsymbol{\lambda}$ with the confusion model. The optimization contains the term $\beta \log(\text{Dir}(\boldsymbol{\lambda}; \alpha))$ and the $Q$-difference (\ref{eq:q_diff}) is:
\begin{eqnarray}
\lefteqn{Q(q^{(j)}, \boldsymbol{\mu}^{(j)}+\boldsymbol{\delta}) - Q(q^{(j)}, \boldsymbol{\mu}^{(j)})+}\nonumber\\
&& \beta \log(\text{Dir}(e^{\boldsymbol{\mu}^{(j)}+\boldsymbol{\delta}}/Z_1)) - \beta \log(\text{Dir}(e^{\boldsymbol{\mu}^{(j)}}/Z_0))\nonumber\\
\label{eq:q_diff_map}
\end{eqnarray}
where $Z_0, Z_1$ are the corresponding normalizing constants. In order to obtain a lower bound on the difference in the second line
of (\ref{eq:q_diff_map}), we consider the following chain of equalities:
\begin{eqnarray}
\lefteqn{\log(\text{Dir}(e^{\boldsymbol{\mu}^{(j)}+\boldsymbol{\delta}}/Z_1)) - \log(\text{Dir}(e^{\boldsymbol{\mu}^{(j)}}/Z_0)) =}\nonumber\\
&&\mspace{-30mu}=(\alpha-1)\sum_t \log\left(\frac{e^{\mu_t^{(j)} + \delta_t}/e^{\mu_t^{(j)}}}{\left(\sum_{t^\prime} e^{\mu_{t^\prime}^{(j)} + \delta_{t^\prime}}\right)/\left(\sum_{t^\prime} e^{\mu_t^{(j)}}\right)}\right)\nonumber\\
&&\mspace{-30mu}=(\alpha-1)\left[\sum_t \delta_t - T \log\left(\sum_{t} \frac{e^{\mu_{t}^{(j)}}}{\sum_{t^\prime} e^{\mu_{t^\prime}^{(j)}}} e^{\delta_t}\right)\right]
\label{eq:prior_diff1}
\end{eqnarray}
We distinguish two cases:\\
(i) {\bf Case 1:} $\alpha < 1$.\\
We 
apply Jensen's inequality to obtain the following lower bound on (\ref{eq:prior_diff1}),
\begin{eqnarray}
&&(\alpha-1)\left[\sum_t \delta_t - T \sum_t \frac{e^{\mu_{t}^{(j)}}}{\sum_{t^\prime} e^{\mu_{t^\prime}^{(j)}}} \delta_t\right]\nonumber\\
&&=(\alpha-1)\left[\sum_t \delta_t - T \sum_t \lambda_t^{(j)} \delta_t\right]~.
\label{eq:prior_diff2}
\end{eqnarray}
The lower bound (\ref{eq:prior_diff2}) now becomes part of the lower bound in (\ref{eq:q_diff_lb}) (for the case of the 1-best) or the lower bound (\ref{eq:q_diff_lb_ec}) (for the case of expected counts), and after differentiating with respect to
 $\delta_t$ and setting the result equal to zero we obtain the update equation:
\begin{eqnarray}
\lefteqn{\mspace{-0mu}e^{\mu_{t}^{(j)}+\delta_t} = (A + \beta(\alpha-1)(1-T \lambda_t^{(j)})) \times}\nonumber\\
&&\Bigg(\sum_{i=1}^M \frac{\sum_{w \in b_i} q(w|t)}{\sum_{t^\prime} e^{\mu_{t^\prime}^{(j)}} \sum_{w \in b_i} q(w|t^\prime)}\Bigg)^{\!\!-1}
\label{eq:update2}
\end{eqnarray}
where $A$ in (\ref{eq:update2}) is equal to
\begin{equation}
\sum_{i=1}^M \sum_{w \in b_i} \frac{r_i^{(j)}(w | \widehat{w}_i) e^{\mu_{t}^{(j)}} q(w|t)}{\sum_{t^\prime} e^{\mu_{t^\prime}^{(j)}}q(w|t^\prime)}
\label{eq:A1}
\end{equation}
in the case of using just the 1-best, and
\begin{equation}
\sum_{i=1}^M \sum_{w,w^\prime \in b_i} \frac{r_i^{(j)}(w | w^\prime) s_i(w^\prime) e^{\mu_{t}^{(j)}} q(w|t)}{\sum_{t^\prime} e^{\mu_{t^\prime}^{(j)}}q(w|t^\prime)}~.
\label{eq:A2}
\end{equation}
in the case of using the lattice.\\
(ii) {\bf Case 2:} $\alpha \geq 1$.\\
We apply the well-known inequality $\log(x) \leq x-1$ and obtain the following lower bound on (\ref{eq:prior_diff1}),
\begin{equation}
(\alpha-1)\left[\sum_t \delta_t - T\left(\sum_t \frac{e^{\mu_{t}^{(j)}+\delta_t}}{\sum_{t^\prime} e^{\mu_{t^\prime}^{(j)}}}-1\right)\right]
\label{eq:prior_diff3}
\end{equation}
As before, the lower bound (\ref{eq:prior_diff3}) now becomes part of the lower bound in (\ref{eq:q_diff_lb}) (for the case of the 1-best) or the lower bound (\ref{eq:q_diff_lb_ec}) (for the case of expected counts), and after differentiating with respect to
 $\delta_t$ and setting the result equal to zero we obtain the update equation:
\begin{eqnarray}
\lefteqn{\mspace{-0mu}e^{\mu_{t}^{(j)}+\delta_t} = (A+\beta(\alpha-1)) \times}\nonumber\\
&&\mspace{-40mu}\times\Bigg(\sum_{i=1}^M \frac{\sum_{w \in b_i} q(w|t)}{\sum_{t^\prime} e^{\mu_{t^\prime}^{(j)}} \sum_{w \in b_i} q(w|t^\prime)}-\frac{\beta(\alpha-1)T}{\sum_{t^\prime} e^{\mu_{t^\prime}^{(j)}}}\Bigg)^{\!\!-1}
\label{eq:update3}
\end{eqnarray}
where $A$ in (\ref{eq:update3}) is equal to (\ref{eq:A1}) or (\ref{eq:A2}), depending on whether the log-likelihood of the 1-best or the expected log-likelihood is maximized.

\begin{figure*}[tb]
\begin{minipage}{\textwidth}
\center
\includegraphics[width=.3\textwidth]{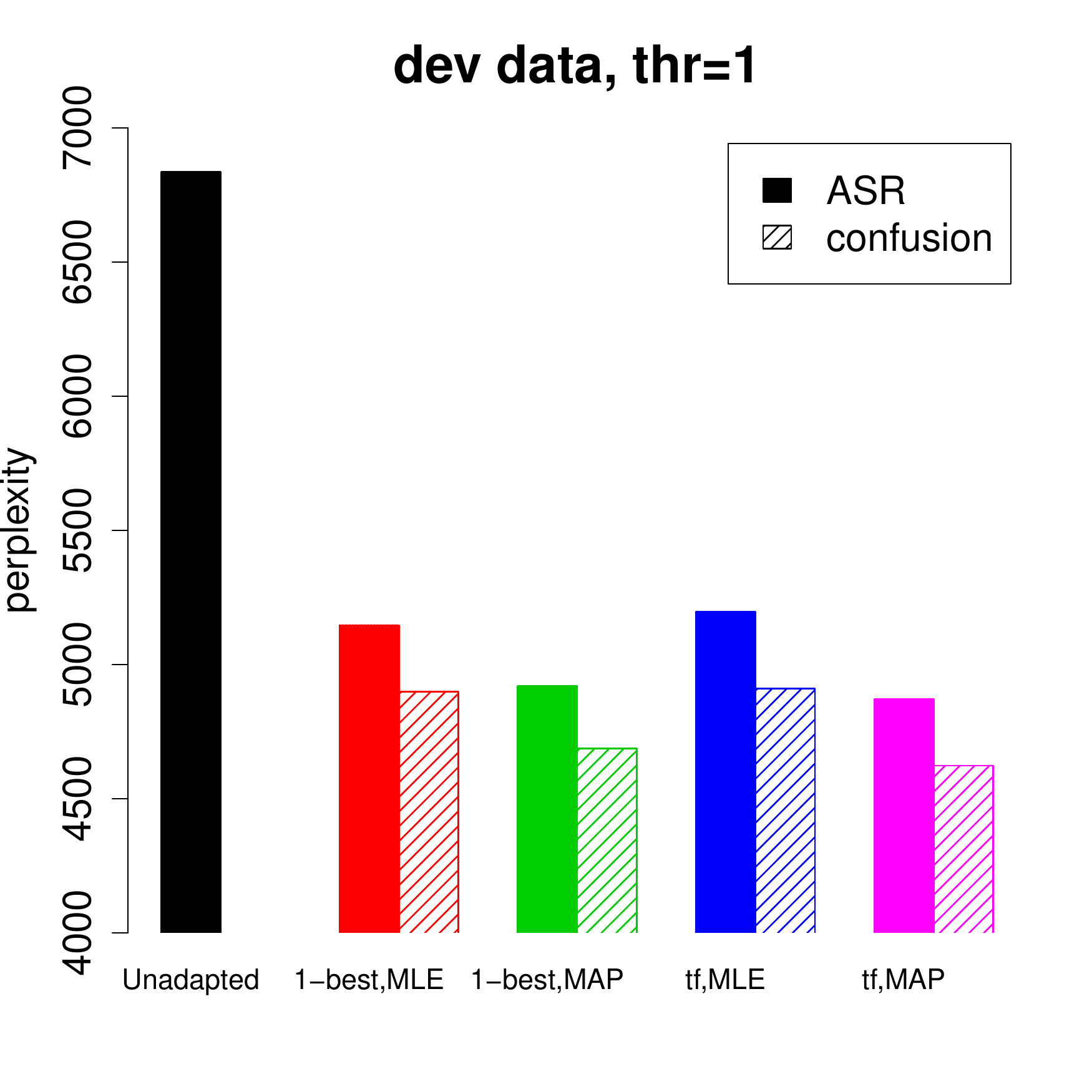}\includegraphics[width=.3\textwidth]{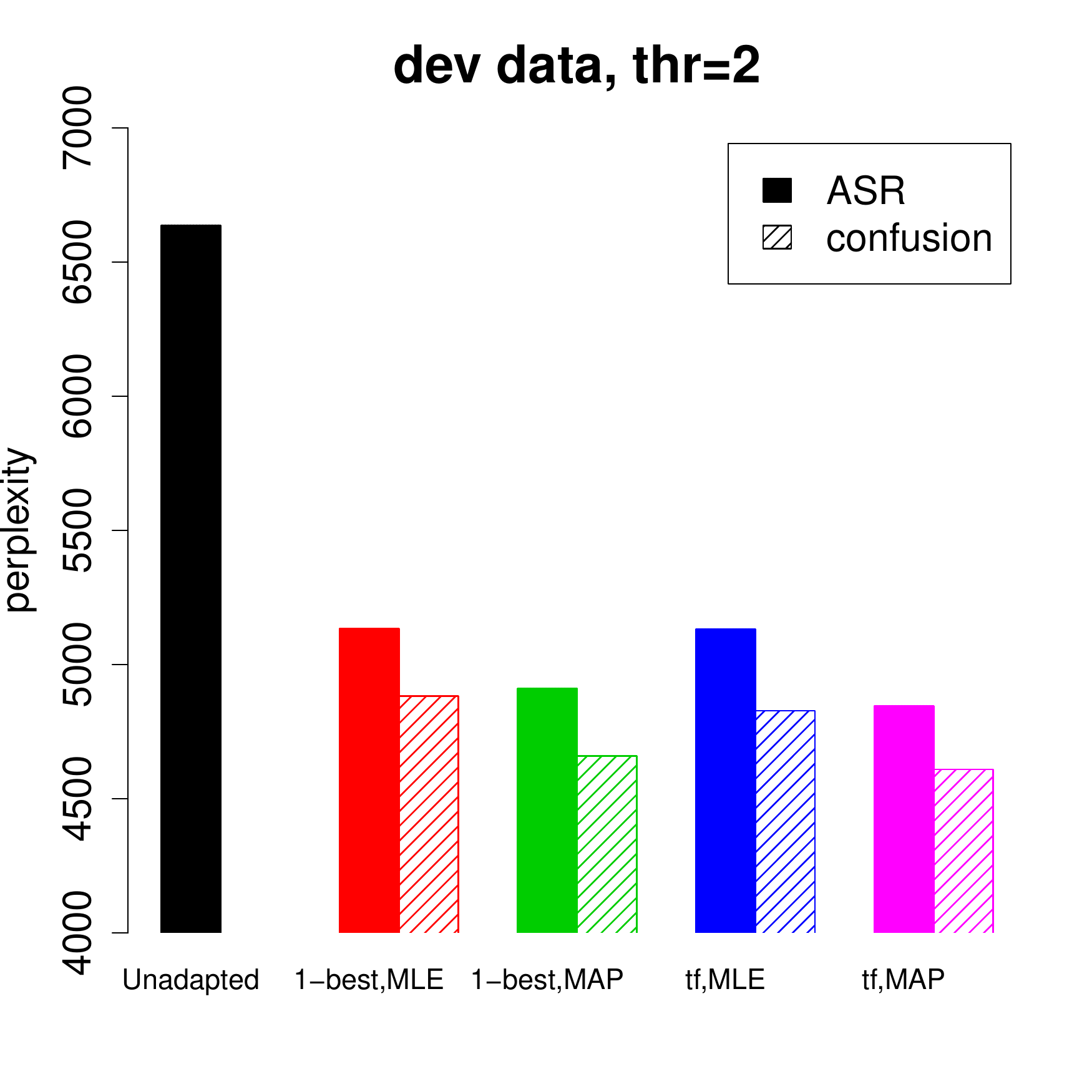}\includegraphics[width=.3\textwidth]{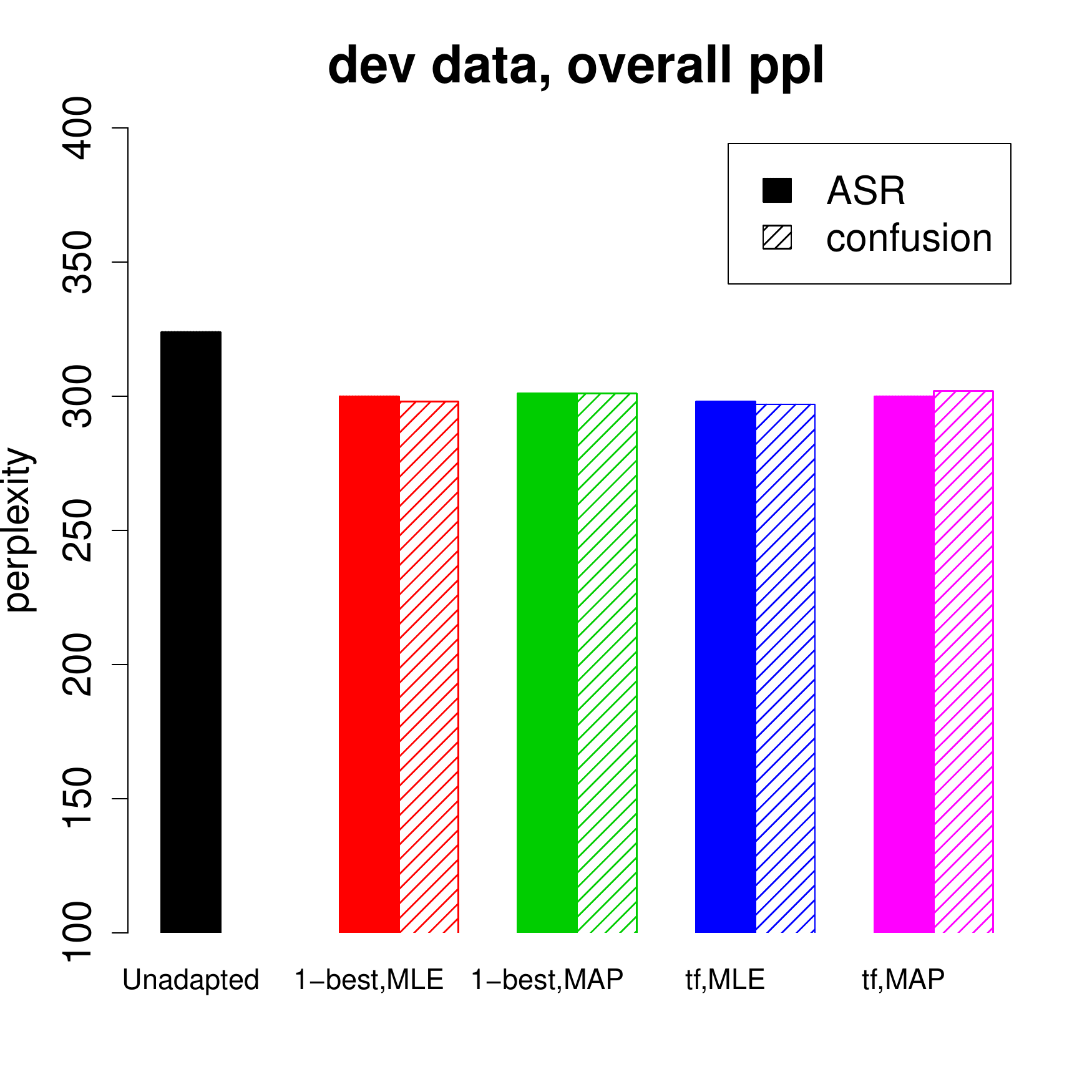} \\
\vspace{-.5cm}
\includegraphics[width=.3\textwidth]{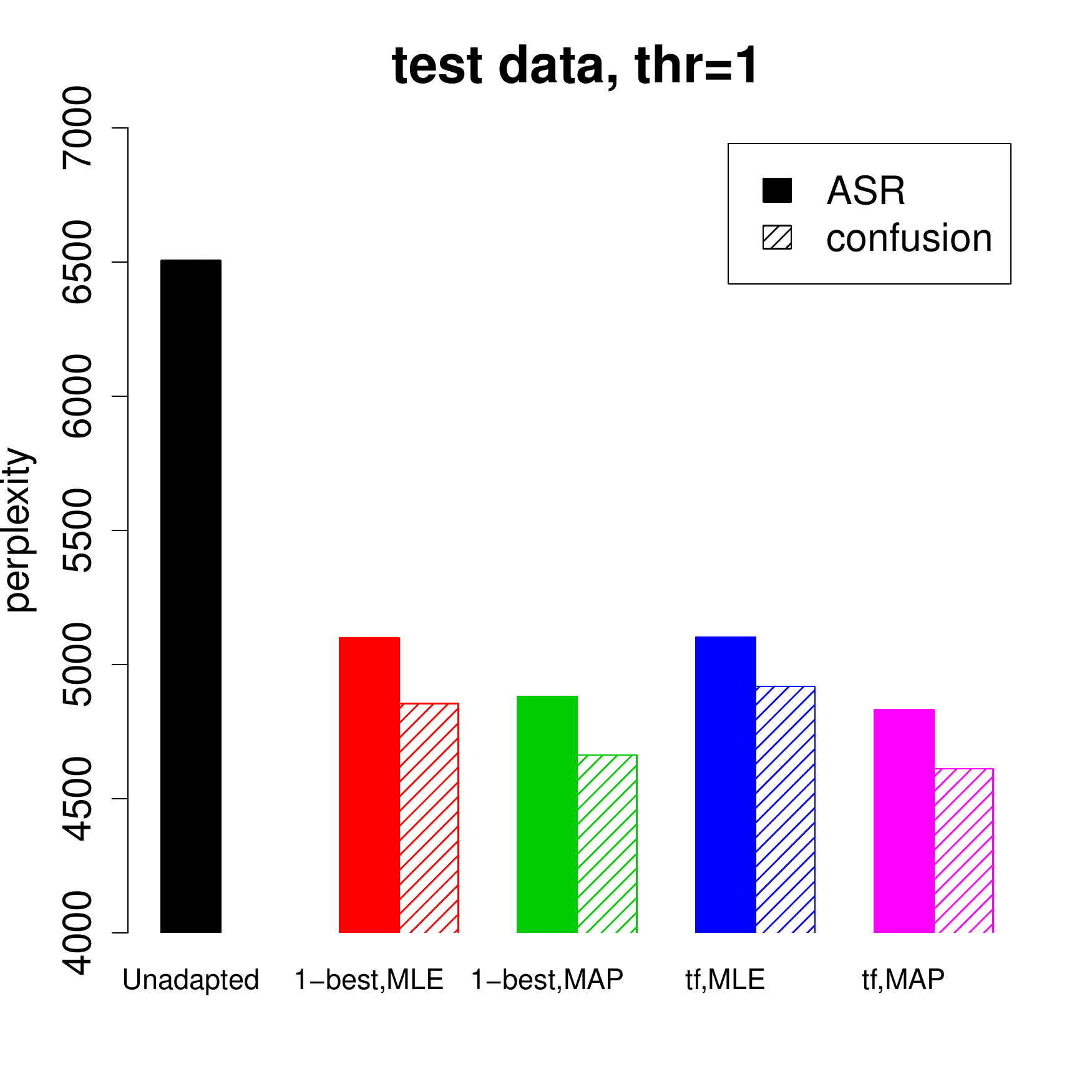}\includegraphics[width=.3\textwidth]{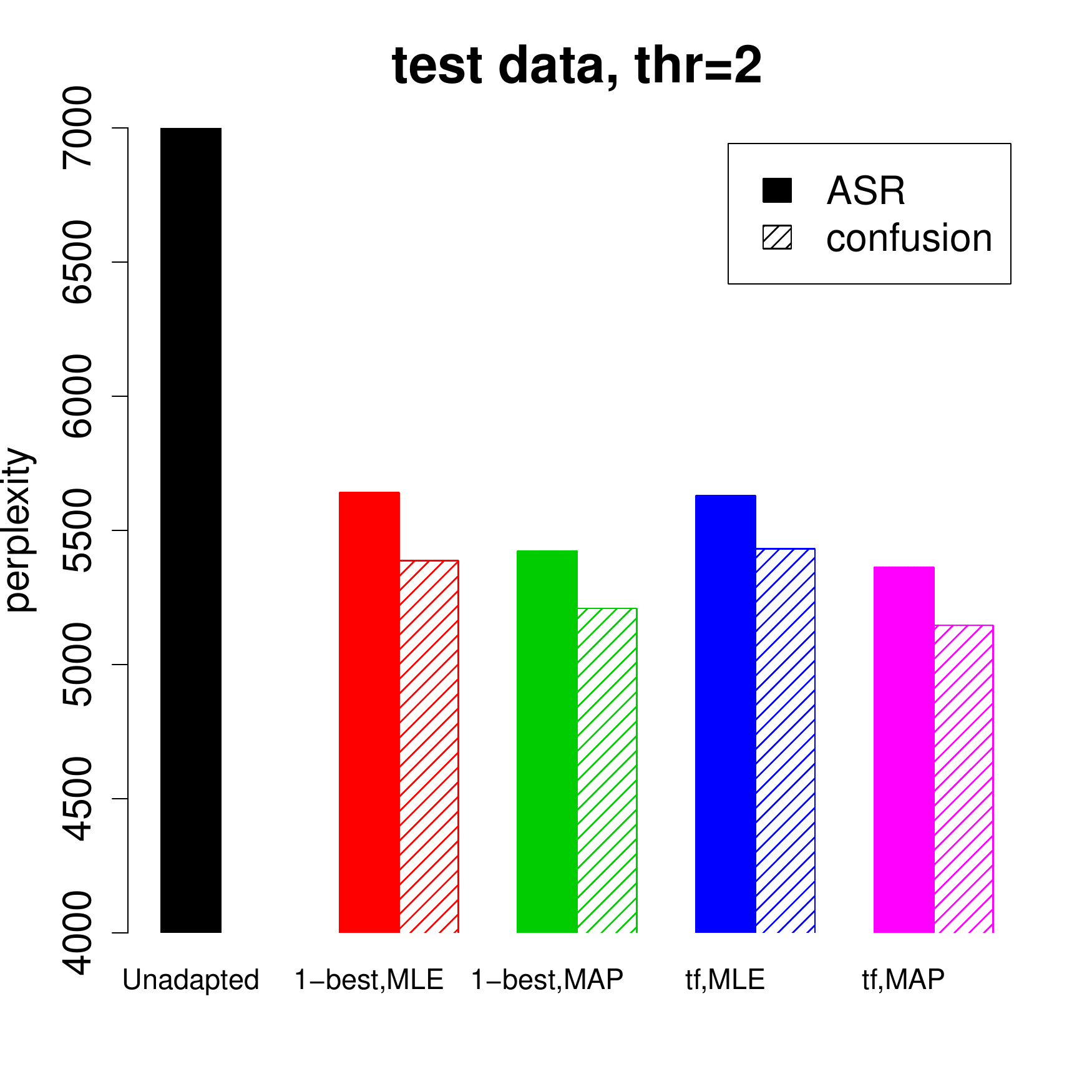}\includegraphics[width=.3\textwidth]{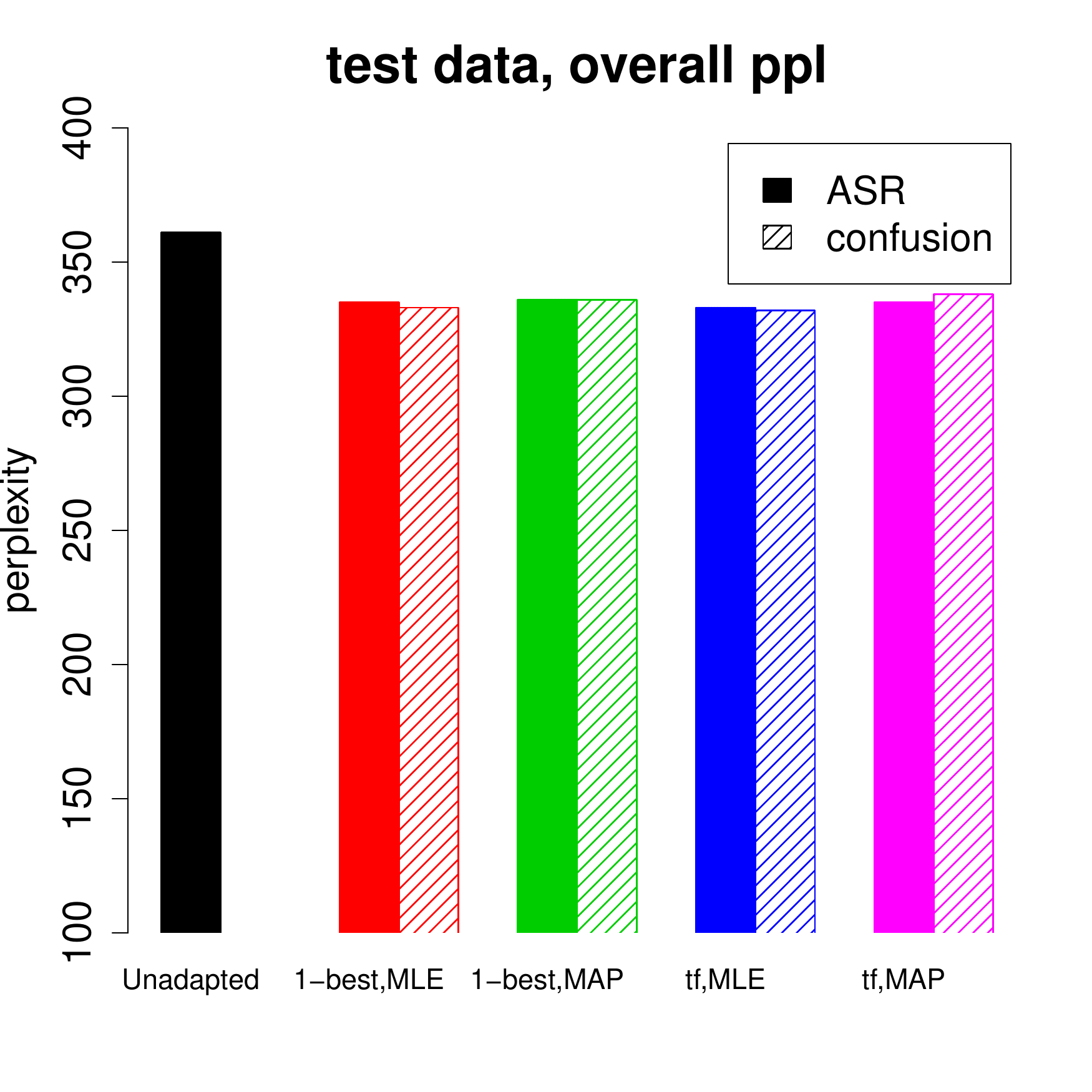}
\end{minipage}
\caption{Perplexities of each language model. ``Content-word'' perplexity is computed over those words whose reference transcript counts do not exceed the threshold {\tt thr} (shown at the top of each plot). We consider MLE and MAP solutions based on 1-best and expected counts ($\tf$) for both ASR output self-training (solid bars) and our confusion model (dashed bars). Overall perplexity (which is dominated by function words) is not affected significantly.}
\label{fig:ppl}
\end{figure*}

\paragraph{Summary:} We summarize our methods as graphical models in Figure \ref{fig:graphical_model}.
We estimate the topic distributions $\boldsymbol{\lambda}^{\cal{C}}$ for a conversation $\cal{C}$
by either self-training directly on the ASR output, or including our ASR channel confusion model. For training on the ASR output,
we rely on MLE 1-best training ((\ref{eq:posterior1}),(\ref{eq:lambda_update_1}), left figure), or  
expected counts from the lattices (\ref{eq:lambda_update_2}). For both settings we also consider MAP estimation: 1-best (\ref{eq:lambda_map_1}) and expected counts (\ref{eq:lambda_map_2}). When using the ASR channel confusion model, we derived parallel cases
for MLE 1-best ((\ref{eq:update1}), middle figure) and expected counts (\ref{eq:update1_part2}), as well as the MAP (right figure) training of each ((\ref{eq:A1}),(\ref{eq:A2})).

\section{Experimental Results}
\label{sec:experiments}
We compare self-training, as described in Section \ref{sec:self_supervised}, with our confusion approach, as described in Section \ref{sec:asr_channel_approach}, on topic-based language model adaptation.

\paragraph{Setup}
Speech data is taken from the Fisher telephone conversation speech corpus, which has been split into 4 parts: set $A_1$ is one hour of 
speech used for training the ASR system (acoustic modeling). We chose a small training set to simulate a high WER condition (approx. 56\%) since we are primarily interested in low resource settings. While conversational speech is a challenging task even with many hours of training data, we are interested in settings with a tiny amount of training data, such as for new domains, languages or noise conditions.
Set
$A_2$, a superset of $A_1$, contains 5.5mil words of manual transcripts, used for training the topic-based distribution $q(\cdot|t)$.
Conversations in the Fisher corpus are labeled with 40 topics so we create 40 topic-based unigram distributions. These are smoothed based on 
the vocabulary of the recognizer using the Witten-Bell algorithm \cite{CG96}.
Set B, which consists of 50 hours and is disjoint from the other sets, is used as a 
development corpus for tuning the MAP parameter $\beta(\alpha-1)$.
Finally, set C (44 hours) is used as a blind test set. The ASR channel and the topic proportions $\boldsymbol{\lambda}$ are learned in
an unsupervised manner on both sets B and C.
The results are reported on 
approximately 5-hour subsets of sets B and C, consisting of 35 conversations each.

BBN's ASR system, Byblos, was used in all ASR experiments. It is a multi-pass LVCSR system that uses state-clustered
Gaussian tied-mixture models at the triphone and quinphone levels \cite{prasad_et_al05}. The audio features are transformed using cepstral normalization, HLDA and VTLN. Only ML estimation was used. Decoding performs three passes: a forward and backward
pass with a triphone acoustic model (AM) and a 3-gram language model (LM), and rescoring using quinphone AM and a 4-gram LM. These three steps are repeated after speaker adaptation using CMLLR.
The vocabulary of the recognizer is 75k words. References of the dev and test sets have vocabularies of 13k and 11k respectively.

\remove
{
Perplexity is given as:
\begin{equation}
\text{PPL}(p) \triangleq 10^{-\frac{1}{|C|} \sum_{i=1}^{|C|} \log_{10}(p(w_i))},
\label{eq:ppl}
\end{equation}
where $p$ represents the estimated language model, and $|C|$ is the size of the (dev or test) corpus.
}

\paragraph{Content Words}
Our focus on estimating confusions suggests that improvements would be manifest for content words, as opposed to frequently occurring function words. This would be a highly desirable improvement as more accurate content words
lead to improved readability or performance on downstream tasks, such as information retrieval and spoken term detection. 
As a result, we care more about reducing perplexity on these content words than reducing overall scores, which give too much emphasis to function words, the most frequent tokens in the reference transcriptions. 

To measure content word (low-frequency words) improvements we use the method of \newcite{Wu:2000fk}, who compute a constrained version of perplexity focused on content words. We restrict the computation to only those words whose counts in the reference transcripts are at most equal to a threshold {\em thr}.
Perplexity \cite{Jelinek97} is measured on the manual transcripts of both dev and test data based on the formula $\text{PPL}(p) \triangleq 10^{-\frac{1}{|C|} \sum_{i=1}^{|C|} \log_{10}(p(w_i))}$, where $p$ represents the estimated language model, and $|C|$ is the size of the (dev or test) corpus.
We emphasize that constrained perplexity is not an evaluation metric, and directly optimizing it would foolishly hurt overall perplexity. However, if overall perplexity remains unchanged, then improvements in content word perplexity reflect a shift of the probability mass, emphasizing corrections in content words over the accuracy of function words, a sensible choice for improving output quality.

\paragraph{Results}
First we observe that overall perplexity (far right of Figure \ref{fig:ppl}) remains unchanged; none of the differences between models with and without confusion estimates are statistically significant. However, as expected, the confusion estimates significantly improves the performance on content words (left and center of Figure \ref{fig:ppl}.) The confusion model gives modest (4-6\% relative) but statistically significant ($p<2\cdot10^{-2}$) gains \emph{in all conditions} for content (low-frequency) word. Additionally, the MAP variant (which was tuned 
based on low-frequency words) gives gains 
over the MLE version \emph{in all conditions}, for both the self-supervised and the confusion model cases. 
This indicates that modeling confusions focuses improvements on content words, which improve readability and downstream applications.

\paragraph{ASR Improvements}
Finally, we consider how our adapted language models can improve WER of the ASR system. We used each of the language models with the recognizer to produce transcripts of the dev and test sets. Overall, the best language model (including the confusion model) yielded no change in the overall WER (as we observed with perplexity). However, in a rescoring experiment, the adapted model with confusion estimates resulted in a 0.3\% improvement in content WER (errors restricted to words that appear at most 3 times) over the unadapted model, and a 0.1\% improvement over the regular adapted model. This confirms that our improved language models yield better recognition output, focused on improvements in content words.

\section{Conclusion}
We have presented a new model that captures the confusions (errors) of the ASR channel. When incorporated with adaptation of a topic-based language model, we observe improvements in modeling of content words that improve readability and downstream applications. Our improvements are consistent across a number of settings, including 1-best and lattice self-training on conversational speech. Beyond improvements to language modeling, we believe that our confusion model can aid other speech tasks, such as topic classification. We plan to investigate other tasks, as well as better confusion models, in future work.

\bibliographystyle{acl2012}

\end{document}